\documentclass{IEEEtran}
\usepackage{cite}
\usepackage{amsmath,amssymb,amsfonts}
\usepackage{algorithmic}
\usepackage{graphicx}
\usepackage[colorlinks,linkcolor=red]{hyperref}
\usepackage{textcomp}
\usepackage{verbatim}
\usepackage{multirow}
\usepackage{bm}
\usepackage{svg}
\usepackage{color}
\def\BibTeX{{\rm B\kern-.05em{\sc i\kern-.025em b}\kern-.08em
    T\kern-.1667em\lower.7ex\hbox{E}\kern-.125emX}}
\begin{document}
\title{\huge Remote Sensing Image Scene Classification with Self-Supervised Paradigm under Limited Labeled Samples}
\author{Chao Tao \IEEEmembership{Member, IEEE}, Ji Qi, Weipeng Lu, Hao Wang and Haifeng Li\^*, \IEEEmembership{Member, IEEE}
\thanks{This work was supported by National key research and development projects (grant number 2018YFB0504500), National Natural Science Foundation of China (grant number 41771458, 41301453), Young Elite Scientists Sponsorship Program by Hunan Province of China under Grant 2018RS3012, and Hunan Science and Technology Department Innovation Platform Open Fund Project under Grant 18K005.}
\thanks{C. Tao, J. Qi, W.P. Lu, H. Wang and H.F. Li are with
the School of Geosciences and Info-Physics, Central South University,
Changsha 410083, China (Corresponding author: H.F. Li, E-mail: lihaifeng@csu.edu.cn).}}

\maketitle

\begin{abstract}
    With the development of deep learning, supervised learning methods perform well in remote sensing images (RSIs) scene classification. However, supervised learning requires a huge number of annotated data for training. When labeled samples are not sufficient, the most common solution is to fine-tune the pre-training models using a large natural image dataset (e.g. ImageNet). However, this learning paradigm is not a panacea, especially when the target remote sensing images (e.g. multispectral and hyperspectral data) have different imaging mechanisms from RGB natural images. To solve this problem, we introduce a new self-supervised learning (SSL) mechanism to obtain the high-performance pre-training model for RSIs scene classification from large unlabeled data. Experiments on three commonly used RSIs scene classification dataset demonstrated that this new learning paradigm outperforms the traditional dominant ImageNet pre-trained model. Moreover, we analyze the impacts of several factors in SSL on RSIs scene classification task, including the choice of self-supervised signals, the domain difference between source and target dataset, and the amount of pre-training data. The insights distilled from our studies can help to foster the development of SSL in remote sensing community. Since SSL could learn from unlabeled massive RSIs which are extremely easy to obtain, it will be a potentially promising way to alleviate dependence on labeled samples and thus efficiently solve many problems, such as global mapping.
\end{abstract}


\section{Introduction}
\label{sec:introduction}
\IEEEPARstart{R}{emote} sensing technologies are playing an increasingly important role in global observing missions due to the wide range of observations and high temporal resolution. Particularly, RSIs scene classification, which aims to classify scene images into different semantic categories, has been a hot topic driven by applications such as land resource management and urban planning\cite{RN259,RN13}.

To achieve accurate scene classification, how to extract discriminative features from RSIs to precisely represent the semantic content of scene has attracted wide attention. In recent years, with the powerful hierarchical feature extraction capabilities of deep convolutional neural networks (DCNNs), deep learning based methods have made significant progresses in RSIs scene understanding\cite{RN245,RN13,RN394}. However, learning DCNNS generally requires large datasets, and building a big remote sensing dataset like ImageNet (with more than 10 million annotated natural image samples) is almost impossible, as the accurate annotation of RSIs is tedious work requiring rich experience and sound geographic knowledge. Pre-training methods  can solve such problem effectively. These methods pre-train the DCNNs on  a large labeled RGB natural image dataset (source data) and then fine-tune the network with small remote sensing data (target data). Although many researches\cite{RN70,RN68,RN67} have demonstrated that the features extracted from ImageNet pre-trained DCNNs can generalize well to aerial image scene classification tasks, pre-training CNNs on ImageNet has two limitations: 1) It may provide no benefit if there exists significant domain difference between source and target datasets. 2) Fine-tuning cannot be applied directly if the data types and imaging mechanisms of source and target dataset (natural RGB data vs. multispectral data) are quite different.

\begin{figure*}[!t]
\centerline{\includegraphics[width=0.90\textwidth]{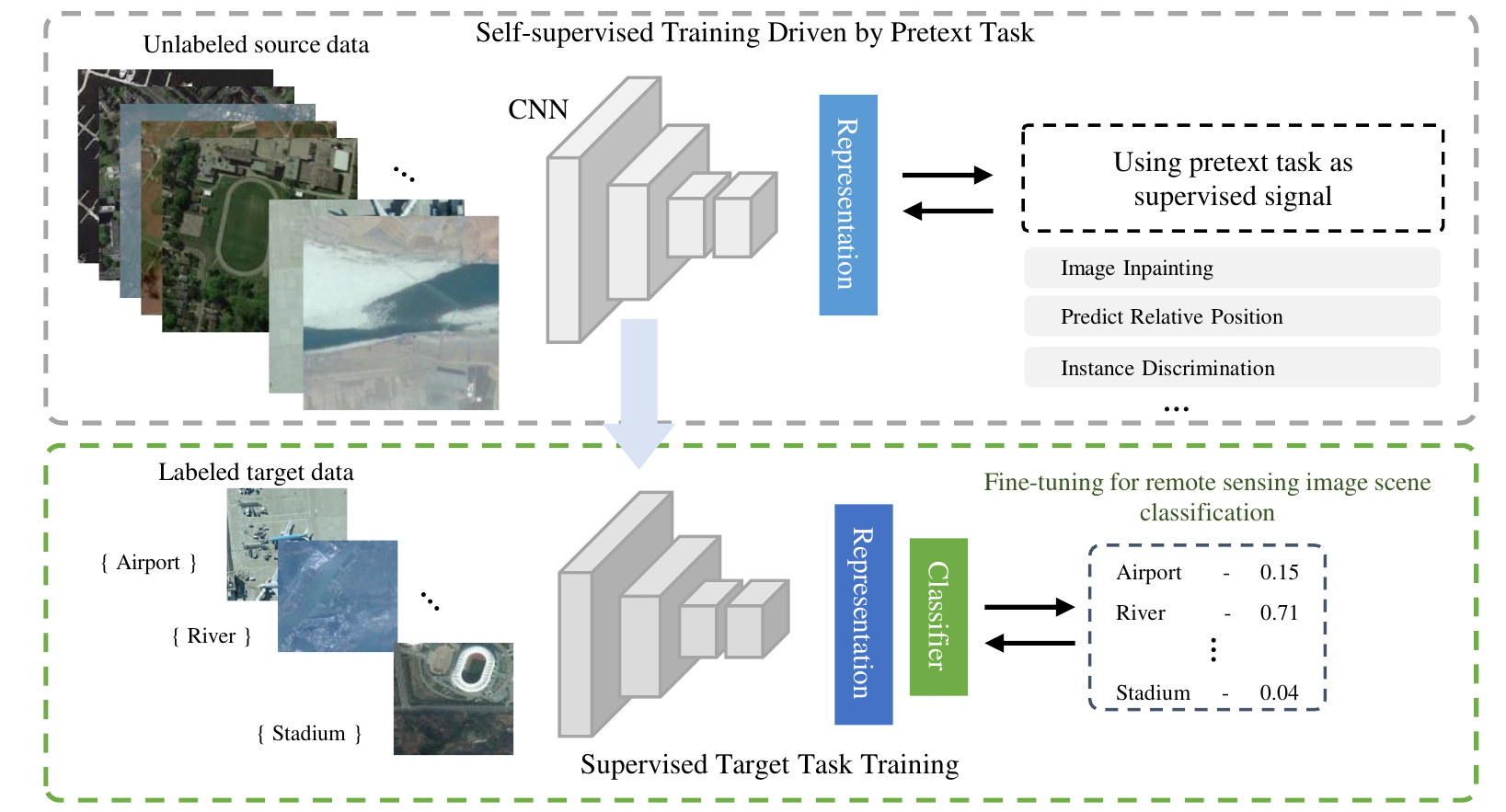}}
\caption{Flowchart of the self-supervised learning paradigm for remote sensing image scene classification.}
\label{fig1}
\end{figure*}

Most recently, a new trend is observed in machine learning, which is learning representations by self-supervised learning (SSL) methods without any additional annotation cost\cite{RN146}. SSL methods can first learn potential useful knowledge from a large amount of unlabeled source data by solving pre-designed tasks (called pretext tasks), then transfer them to target tasks. Inspired by recent advances of SSL in applications like natural language processing\cite{RN143}, nature image classification\cite{RN145} and object detection\cite{RN144}, we believe that this kind of feature learning mechanism is a more effective and robust way for RSIs scene understanding when labeled data is insufficient. The main reasons could be: first, SSL provides a flexible pre-training architecture, because we can use any type of large-scale remote sensing data without human annotation to pre-train DCNNs; second, since we can choose a source dataset similar to the target dataset at low cost for pre-training, the new learning paradigm can potentially alleviate the domain difference and thus ensure the performance of the learned representations on target RSIs scene classification task. Therefore, we introduce the self-supervised feature learning mechanism for RSIs scene classification, and evaluate the feature learning impact of three commonly used pretext tasks on target RSIs scene classification. As far as we know, this is the first time to use the self-supervised learning mechanism in RSIs scene classification. The contributions of this work are mainly in two aspects:

1) We demonstrate that SSL is an entirely new paradigm which learns feature from unlabeled massive images for remote sensing image understanding. This paradigm is extremely suited to RSI understanding tasks because we have very easy access to a large number of RSIs, over different areas and at different times.

2) Experiments on three RSIs scene classification datasets show that the proposed method overpass the traditional dominant ImageNet pre-training approach when labeled data is insufficient.

3) We analyze the effects of several factors on the performance of SSL, which contributes to a deeper understanding of what enables useful self-supervised feature representation for RSIs scene understanding.

\section{Methodology}
\label{sec:methodology}
\subsection{Overview of self-supervised learning paradigm}
In this letter, we suggest a self-supervised learning framework for RSIs scene classification. As shown in Fig. \ref{fig1}, the general pipeline of self-supervised learning paradigm consists of two phases. In the self-supervised training phase, a DCNN is trained to solve predefined pretext tasks for learning potential useful representations on large unlabeled source data. And the learned representations are stored as parameters of the encoder of the DCNN. In the second phase, the learned representations are transferred to target tasks as a pre-trained model. Compared with training from scratch, fine-tuning the pre-trained model with good representations can overcome overfitting and achieve higher performance on target task, especially when labeled samples are insufficient.

\begin{figure}[ht]
\centerline{\includegraphics[width=\columnwidth]{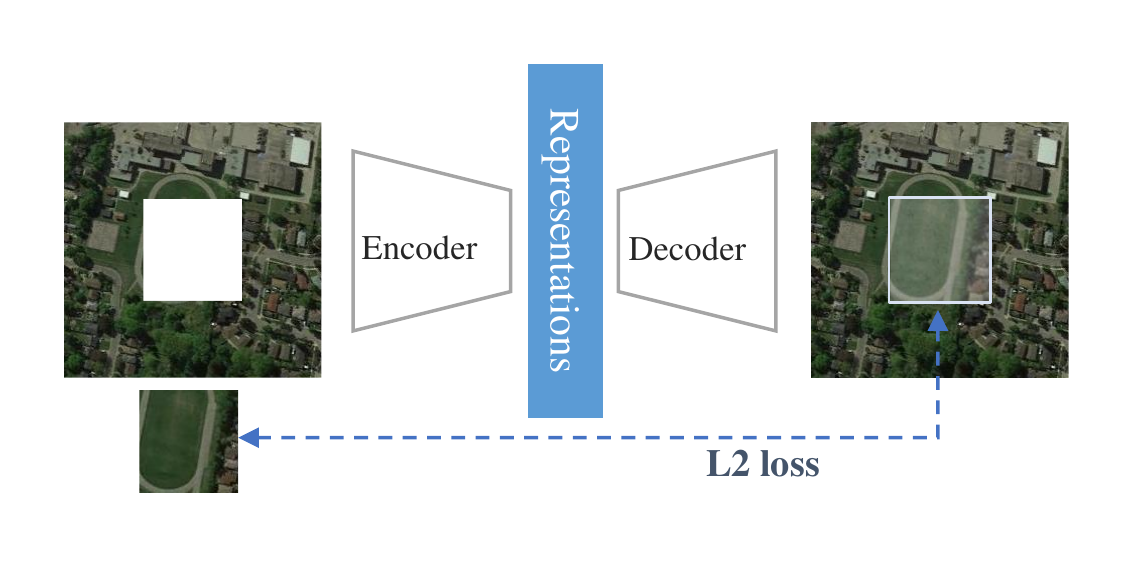}}
\caption{Illustration of image inpainting pretext task. Given the corrupted image (left), the model is used to restore the missing part (right) based on the rest of the image.}
\label{fig2}
\end{figure}

\subsection{Learning representations by solving pretext tasks}
In the self-supervised learning framework, image inpainting\cite{RN116}, predicting relative position\cite{RN148} and instance-wise contrastive learning\cite{RN1} are three common pretext tasks for training DCNN encoders. In the following, we brief the learning algorithms of these three pretext tasks, and evaluate their feature learning impacts on target RSIs scene classification by experiments.

\subsubsection{Image Inpainting}
In image inpainting, model $f(\cdot)$ takes the corrupted image $\tilde{\boldsymbol{x}}_{i}$ as input and is trained to predict the original image. $\tilde{\boldsymbol{x}}_{i}$ is obtained by masking arbitrary regions of ${\boldsymbol{x}}_{i}$. Generally, the objective function for such a task uses L2 loss as shown in \eqref{Eq1}. By optimizing Eq. \eqref{Eq1}, the encoder of DCNN is driven to model pixel-level relationships and local contextual relations within the image for guessing the missing regions based on the rest of the image (Fig. \ref{fig2}).

\begin{equation}
\label{Eq1}
\mathcal{L}_{inpainting}=\left\|f\left(\widetilde{\boldsymbol{x}}_{i}\right)-\boldsymbol{x}_{i}\right\|_{2}^{2}
\end{equation}

\begin{figure}[ht]
\centerline{\includegraphics[width=0.97\columnwidth]{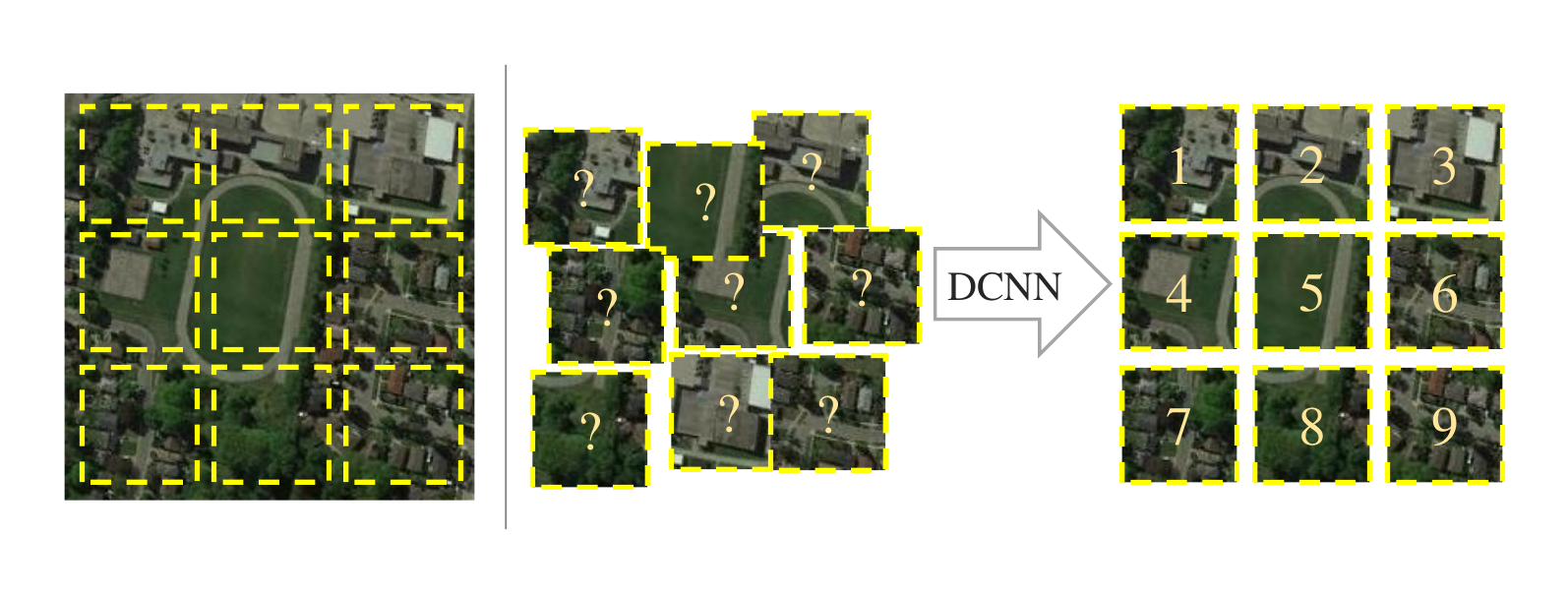}}
\caption{A typical example for predicting the relative position: $3 \times 3$ Jigsaw puzzles. A DCNN takes nine patches as input and predict their positions.}
\label{fig3}
\end{figure}

\subsubsection{Predict the Relative Position}
Image parts have rich complex spatial or sequential relations, especially natural images. For instance, in portrait photos the head is above the body. Therefore, various models regard recognizing relative positions between parts of images as the pretext task for self-supervised learning. The relative positions could be between two patches from a sample\cite{RN147}, or between shuffled segments of an image (solve jigsaw)\cite{RN148}, as shown in Fig. \ref{fig3}. Given an image meshed into $m \times n$ patches $\boldsymbol{p}_{i,j} (i=1,2,\ldots,m \text { and } j=1,2,\ldots,n)$, model $f(\cdot)$ learns contextual relationships between patches by optimizing the loss function in Eq. \eqref{Eq2}:
\begin{equation}
\label{Eq2}
\mathcal{L}_{jigsaw}=-\sum_{i=1}^{m} \sum_{j=1}^{n} P_{i, j} \log \left(f\left(\boldsymbol{p}_{i, j}\right)\right),
\end{equation}
where $P_{i,j}$ is the ground truth position of $\boldsymbol{p}_{i,j}$.

\subsubsection{Instance Discrimination}
\begin{figure}[!t]

\centerline{\includegraphics[width=\columnwidth]{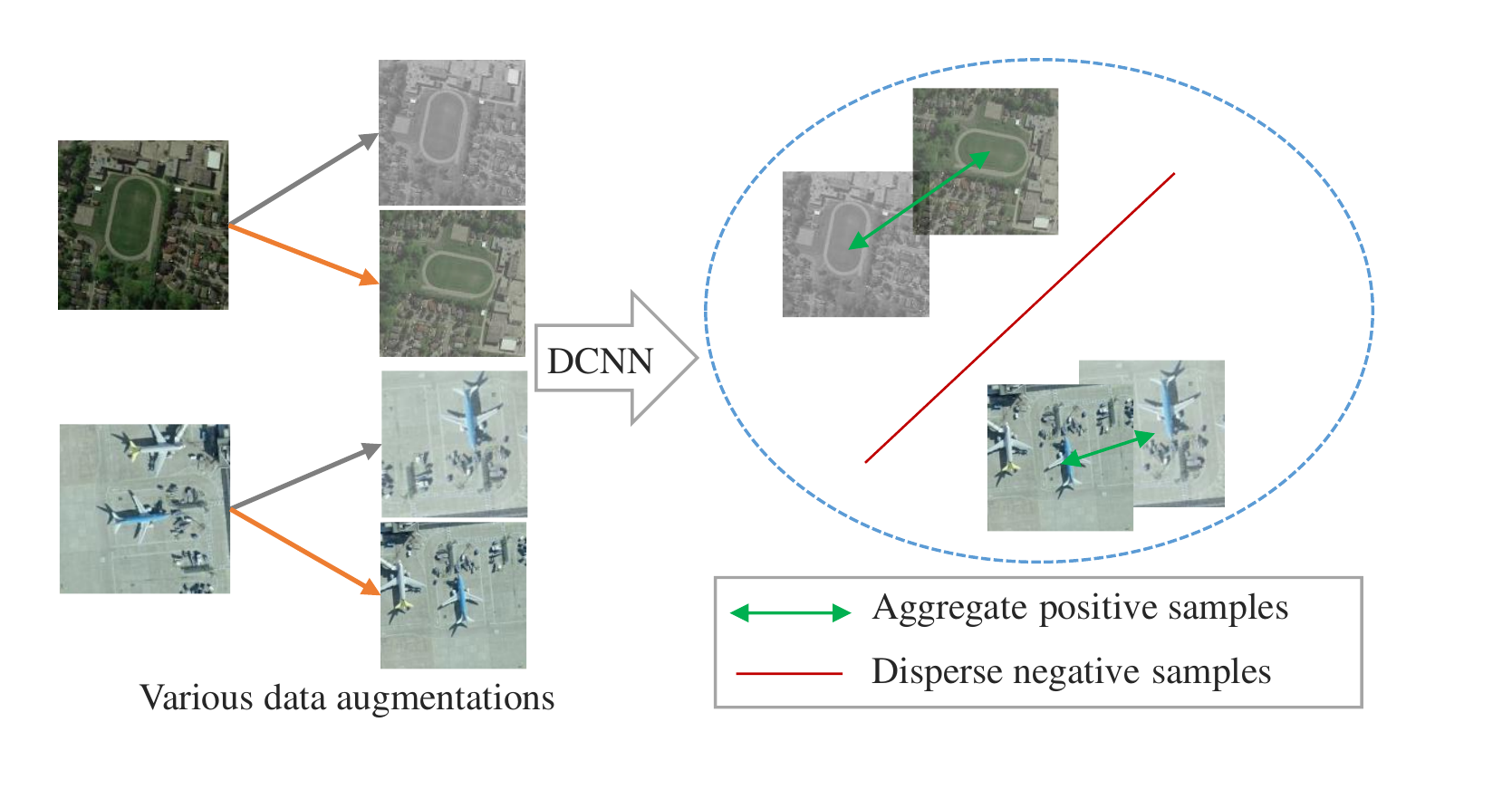}}
\caption{Illustration of the instance discrimination pretext task. A DCNN is asked to draw near multiple augmentation views of an image sample, and pull away  one sample from the other samples by embedding them to a proper feature space.}
\label{fig4}
\end{figure}

Different from the above pretext tasks, instance discrimination (ID, or instance-wise contrastive learning) tasks classify examples as their own labels\cite{RN144,RN145}. Specifically, the ID-based self-supervised learning (IDSSL) method takes different augmented views of a sample as positive samples, and takes other different samples as negative ones. Then, a DCNN is trained to distinguish between positive and negative samples by embedding them to a proper feature space with learned representation $f(\cdot)$ (Fig. \ref{fig4}). A minibatch of $N$ samples is augmented to be $2N$ samples $\tilde{\boldsymbol{x}}_{i} (i=1,2,\ldots,2N)$. For a pair of positive samples $(\tilde{\boldsymbol{x}}_{i}, \tilde{\boldsymbol{x}}_{j})$, other $2(N-1)$ samples are negative ones. Then, a pairwise contrastive loss, the NT-Xent loss\cite{RN145}, is defined as Eq. \eqref{Eq3}:
\begin{equation}
\label{Eq3}
\ell_{i, j}=-\log \frac{\exp \left(\operatorname{sim}\left(f\left(\hat{\boldsymbol{x}}_{i}\right), f\left(\hat{\boldsymbol{x}}_{j}\right)\right) / \tau\right)}{\sum_{k=1}^{2 N} \mathbb{I}_{[k \neq i]} \exp \left(\operatorname{sim}\left(f\left(\hat{\boldsymbol{x}}_{i}\right), f\left(\hat{\boldsymbol{x}}_{k}\right)\right) / \tau\right)},
\end{equation}
where $\tau$ denotes a temperature parameter, and $\mathbb{I}_{[k \neq i]}\in{\{0,1\}}$ is an indicator function evaluating to $1 \operatorname{iff} [k \neq i]$. The similarity measure function $\operatorname{sim}(\boldsymbol{u}, \boldsymbol{v})=\boldsymbol{u}^{\mathrm{T}} \boldsymbol{v} /\|\boldsymbol{u}\|\|\boldsymbol{v}\|$ denotes the cosine similarity between vectors $\boldsymbol{u}$ and $\boldsymbol{v}$. $\ell_{i, j}$ is asymmetrical, so the total loss of a minibatch can be computed by Eq. \eqref{Eq4}:
\begin{equation}
\label{Eq4}
\mathcal{L}=\frac{1}{2 N} \sum_{k=1}^{N}\left[\ell_{2 i-1,2 i}+\ell_{2 i, 2 i-1}\right].
\end{equation}

\section{Experiment}
\subsection{Datasets Description and Experiment Designing}
The scene classification experiments used three public datasets with few labeled samples, which are EuroSAT, AID and NR. These datasets can be divided into low-resolution multi-spectral RSIs datasets and high-resolution RSIs datasets:

1) Multi-spectral RSIs datasets

{\bf EuroSAT}\cite{RN273} contains 27,000 samples of 10 categories, collected from Sentinel-2 satellite. These images in 13 bands have a spatial resolution of 30-10m. We used all the 27,000 samples without annotation as the source data for SSL pre-training, and 5,400 samples were used for test.

2) High-resolution RSIs datasets

{\bf Aerial Image dataset} (AID)\cite{RN5} contains 10,000 samples of 30 classes, collected form Google Earth. These overhead scene images in RGB color space have a resolution of approximately 8-0.5m. We used all the 10,000 samples without annotation for SSL pre-training, and 8,000 for testing.

{\bf NWPU-RESISC45 dataset} (NR)\cite{RN2} contains 31,500 samples of 45 scene categories, collected from Google Earth. The spatial resolution varies from about 30m to 0.2m per pixel for most images. We used all the 31,500 samples without annotation for SSL pre-training, and 25,200 for testing.

To evaluate the performance of the SSL-based method on RSIs scene classification task, we carried out the following two experiments in PyTorch environment under the CentOS 7.5 platform with four NVIDIA Tesla V100 (memory 16 GB). The overall accuracy (OA) is used to compare the performance quantitatively.

\begin{table}[!t]
    \centering
    \footnotesize
    
    \caption{The training-testing set configuration of three dataset in the experiments}
    \label{tab01}
    \renewcommand{\arraystretch}{1.3}
    \begin{tabular}{p{16em}ccc}
        \hline
        Dataset                       & EuroSAT   & AID       & NR \\
        \hline
        Classes & 10 & 45 & 30 \\
        Unlabeled samples used for SSL & 21,600 & 25,200 & 8,000 \\
        Labeled samples used for fine-tuning & 5/class & 5/class & 5/class \\
        Samples used for testing & 5,400 & 6,300 & 2,000 \\
        \hline
    \end{tabular}
\end{table}

An overview of the two experiments is as follows:

\begin{itemize}
\item Experiment \uppercase\expandafter{\romannumeral1} aims at analyzing several factors that may affect the pre-training performance of SSL on the target RSIs scene classification task, including the choice of self-supervised signals, the domain difference between source and target dataset, and the amount of pre-training data.
\item Experiment \uppercase\expandafter{\romannumeral2} aims at evaluating the performance of SSL on the task of RSIs scene classification, and demonstrating the advantage of SSL over other methods when labeled training samples are insufficient.
\end{itemize}

\subsection{Experiment I}
In this section, we performed controlled studies to investigate several factors that may affect the pertaining performance of SSL on the target RSIs scene classification task. In the stage of SSL training, we pre-trained ResNet50\cite{RN420} model using the Adam optimizer with a batch size of 256 samples. The learning rate was initially set to 1e-4 and was reduced in a cosine manner within 400 epochs. In the stage of fine-tuning, we used only five labeled samples per category to fine-tune the target task.

\subsubsection{Study of the choice of self-supervised signals}
We evaluated the feature learning performance of different self-supervised signals or pretext tasks on target scene classification task. As shown in TABLE \ref{tab02}, the pretext of instance discrimination consistently outperforms other two pretext tasks by a large margin on all three datasets. These results indicate that choosing an appropriate pretext task is crucial, and the correlation between pretext and target tasks plays an important role in learning representative and transferable features. For solving instance discrimination tasks, models are required to learn high-level abstract features with semantic information, which is crucial for classifying RSIs scenes. While in image inpainting and predicting relative position tasks, models are mainly concerned with the pixel-level relationships and local context.

\begin{table}[!t]
    \centering
    
    \caption{Results on the choice of self-supervised signals for experiment I}
    \label{tab02}
    \renewcommand{\arraystretch}{1.3}
    \begin{tabular}{cccc}
        \hline
        \multirow{2}{*}{Pretext task} & \multicolumn{3}{c}{OA on target scene classification tasks} \\
        \cline{2-4}
                                    & EuroSAT   & AID       & NR \\
        \hline
        Image Inpainting            & 53.81+1.62 & 41.15+1.13 & 17.58+1.22 \\
        Predict Relative Position   & 53.15+1.64 & 50.32+0.79 & 34.23+0.85 \\
        Instance Discrimination     & 76.10+0.27 & 76.80+0.30 & 80.63+0.03 \\
        \hline
    \end{tabular}
\end{table}

\subsubsection{Study of domain difference}
In this study, we compared the performance of models pre-trained using various source datasets by instance discrimination-based SSL on target datasets\footnotemark. TABLE \ref{tab03} shows that pre-training the model using a source dataset similar to the target dataset can make the learned representations suit the target dataset better. For example, source-target pairs with strong domain similarity (e.g., EuroSAT$\to$EuroSAT, AID$\to$AID, and NR$\to$NR) achieves the best results for the three datasets. On the other hand, the performance drops dramatically if the domain difference is large (e.g. from low-resolution multi-spectral RSIs dataset (EuroSAT) to high-resolution RSIs dataset (NR, AID)). This phenomenon exists in most machine learning methods. However, considering SSL training does not need human annotated data, it is efficient and low-cost to obtain source datasets similar to the target dataset.
\footnotetext{Here only RGB bands were used for EuroSAT to ensure that the pre-trained models on the three datasets are transferable to each other datasets.}

\begin{table}[!t]
    \centering
    
    \caption{Results on the domain difference for experiment I}
    \label{tab03}
    \renewcommand{\arraystretch}{1.3}
    \begin{tabular}{cccc}
        \hline
        \multirow{2}{*}{Source data} & \multicolumn{3}{c}{OA on target scene classification tasks for experiment I} \\
        \cline{2-4}
                            & EuroSAT   & AID       & NR \\
        \hline
        EuroSAT (30-10m)    & 76.10+0.27 & 42.80+0.47 & 26.85+0.08 \\
        AID (8-0.5m)	    & 61.84+0.52 & 76.80+0.30 & 54.77+0.15 \\
        NR (30-0.2m)	    & 71.45+0.20 & 74.10+0.07 & 80.63+0.03 \\
        \hline
    \end{tabular}
\end{table}

\subsubsection{Study of the amount of pre-training data}
Since SSL training does not need human annotated data, it is interesting to see whether more pre-training data can bring performance improvements. To this end, given a source training dataset, we created two subsets by randomly sampling 10\% and 50\% of samples and then evaluated the effect of the amount of pre-training data using different subsets of each dataset. The experiment result is shown in TABLE \ref{tab04}. As can be seen, enlarging the training data size generally leads to better performance because deep learning-based method is data hungry. For all three datasets, the performance improved significantly when the amount of pre-training data increases from 10\% to 50\%, but the growth rate gradually slows down when the amount of pre-training data increases from 50\% to 100\%.

\begin{table}[!t]
    \centering
   
    \caption{Results on the amount of unlabeled data for SSL for experiment I}
    \label{tab04}
    \renewcommand{\arraystretch}{1.2}
    \begin{tabular}{ccccc}
        \hline
        \multirow{2}{*}{Source data} & \multirow{2}{*}{Number} & \multicolumn{3}{c}{OA on target scene classification tasks} \\
        \cline{3-5}
         & & EuroSAT   & AID       & NR \\
        \hline
        EuroSAT   & 2,160     & 57.06+0.31 & \textbackslash{} & \textbackslash{} \\
        EuroSAT   & 10,800    & 69.12+0.14 & \textbackslash{} & \textbackslash{} \\
        EuroSAT   & 21,600    & 76.10+0.27 & \textbackslash{} & \textbackslash{} \\
        AID       & 800      & \textbackslash{} & 46.88+0.21 & \textbackslash{} \\
        AID       & 4,000    & \textbackslash{} & 63.80+0.27 & \textbackslash{} \\
        AID       & 8,000    & \textbackslash{} & 76.80+0.30 & \textbackslash{} \\
        NR        & 2,520    & \textbackslash{} & \textbackslash{} & 39.79 \\
        NR        & 12,600   & \textbackslash{} & \textbackslash{} & 68.52  \\
        NR        & 25,200   & \textbackslash{} & \textbackslash{} & 80.63+0.03 \\
        AID and NR          & 33,200  & 72.30+0.15 & 84.20+0.50 & 81.13+0.10 \\
        AID, NR and EuroSAT & 54,800  & 67.69+0.42 & 78.60+0.23 & 75.25+0.24 \\
        \hline
    \end{tabular}
\end{table}

\begin{table*}[!t]
    \centering
    
    \caption{OA of the state-of-the-art methods on three datasets for experiment II}
    \label{tab05}
    \renewcommand{\arraystretch}{1.3}
    \begin{tabular}{ccccccc}
        \hline
        \multirow{3}{*}{Method} & \multicolumn{2}{c}{EuroSAT} & \multicolumn{2}{c}{AID} & \multicolumn{2}{c}{NR} \\ \cline{2-7}
                                & \multicolumn{6}{c}{Number of samples per category} \\ \cline{2-7}
                                & 5 & 20 & 5 & 20 & 5 & 20 \\
        \hline
        ResNet50 from scratch\cite{RN420}   & 53.60+0.58 & 74.14+0.69 & 40.75+1.53 & 59.28+0.88 & 32.70+0.40 & 58.29+1.10 \\
        ImageNet pre-trained ResNet50       & 69.24+0.62 & 80.91+0.28 & 72.77+0.09 & 81.60+0.17 & 59.63+0.06 & 73.74+0.17 \\
        MATAR GANs\cite{RN109}              & 50.92+1.16 & 68.60+0.28 & 53.08+0.44 & 61.90+0.60 & 43.52+0.18 & 59.01+0.24 \\
        IDSSL                               & 76.10+0.27 & 84.68+0.03 & 76.80+0.30 & 80.62+0.22 & 80.63+0.03 & 85.80+0.15 \\
        \hline
    \end{tabular}
\end{table*}

\subsection{Experiment II}
In this section, we compared the instance discrimination-based SSL (IDSSL) method with two techniques, which are (a) ImageNet pre-trained ResNet50 and (b) multiple-layer feature-matching generative adversarial networks (MARTA GANs)\cite{RN109}. Here, MARTA GANs is also an unsupervised representation learning method for RSIs scene classification. It learns features with multiscale spatial information from unlabeled images by proposing a multiple-feature-matching layer combined with GANs. For target tasks, we chose the optimal fine-tuning strategy for each method. Considering the non-negligible domain differences between ImageNet and the above three remote sensing datasets, we fine-tuned the entire ImageNet pre-trained ResNet50. For IDSSL, we fixed the parameters of models' encoder and optimized the full connectivity layer. We used Adam optimizer with a batch size of 64. And the learning rate was initially set to 1e-4 and was reduced in a cosine manner within 200 epochs. Consist with\cite{RN109}, we regarded the model obtained by MARTR GANs as an feature extractor. The extracted features were fed into the SVMs for classification. Since the models pre-trained on ImageNet (natural RGB images) cannot be directly fine-tuned on EuroSAT datasets (multi-spectral RSIs), only RGB channels were used on this dataset for ImageNet pre-trained ResNet50.

From the experimental results in TABLE \ref{tab05}, we got the following findings. First, IDSSL consistently outperforms ImageNet pre-trained model on all three datasets, especially when using only 5 labeled training samples per category. One possible reason is that the domain difference between ImageNet and remote sensing datasets reduces the performance of the ImageNet pre-trained model. In contrast, SSL provides a flexible pre-training architecture, because we can use any type of large-scale remote sensing data with human annotation to pre-train DCNNs. As a result, the new learning paradigm can potentially alleviate the domain difference and ensure the performance of the learned representations on target RSIs scene classification task. Second, IDSSL is less sensitive to the amount of labeled data than other methods. For example, when the fine-tuning data on NR reduces from 20 per class to 5 per class, the performance of IDSSL decreases by only 6.02\%, whereas the relative reduction of ImageNet pre-trained ResNet50 and MATAR GANs are 19.13\% and 26.24\%, respectively. This might be caused by that with robust representations, IDSSL has little risk of overfitting to labels of small-sized target data.

\section{Conclusion and future works}
In this study, we introduce a new learning paradigm, SSL, for RSIs scene classification for the cases of lacking labeled data. Moreover, we performed comprehensive comparative study by analyzing several factors in SSL on RSIs scene classification task and uncovered that the choice of self-supervised signals, the domain difference between source and target dataset, and the amount of pre-training data strongly affect the pre-training performance of SSL. By combining our findings, the SSL based method outperforms traditional dominant ImageNet pre-training approach as well as other state-of-the-art methods by a large margin when labeled data is insufficient. Future work aims at constructing large scale, publicly available benchmarks for different sensors to foster the development of new SSL methods in remote sensing communities and also using the proposed method to applications such as global mapping which struggle with the limited labeled samples and transferability problems.

\bibliographystyle{IEEEtran}
\bibliography{refs}
\end{document}